\pdfoutput=1
\documentclass[11pt]{article}

\usepackage[]{eacl2023}
\usepackage{times}
\usepackage{latexsym}
\usepackage[T1]{fontenc}
\usepackage[utf8]{inputenc}
\usepackage{microtype}
\usepackage{booktabs}
\usepackage{multirow}
\usepackage{makecell}
\usepackage{amsmath}
\usepackage{setspace}
\usepackage{graphicx}
\usepackage{xspace}
\usepackage{CJKutf8}

\usepackage[utf8]{inputenc}
\usepackage{tabularx}
\usepackage{hyperref} 
\usepackage{ruby}
\usepackage{xpinyin}
\usepackage{xspace}
\usepackage{subfigure}

\newcommand{\koreancolor}[1]{\textcolor[HTML]{C16200}{#1}}
\newcommand{\chinesecolor}[1]{\textcolor[HTML]{77AAAD}{#1}}
\newcommand{\korean}[1]{\begin{CJK}{UTF8}{mj}#1\end{CJK}}
  {\list{}{\leftmargin=#1\rightmargin=#1}\item[]}%
  {\endlist}
  
\title{Cross-Cultural Transfer Learning for\\Chinese Offensive Language Detection}

\newcommand{\uestc}{$^1$}
\newcommand{\ku}{$^2$}
\newcommand{\hust}{$^3$}

\author{Li Zhou\uestc \ku, Laura Cabello\ku, Yong Cao\ku\hust, Daniel Hershcovich\ku \\
{\uestc}University of Electronic Science and Technology of China\\
{\ku}Department of Computer Science, University of Copenhagen \\
{\hust}Huazhong University of Science and Technology \\
\texttt{\small li\_zhou@std.uestc.edu.cn, \{lcp,dh\}@di.ku.dk, yongcao\_epic@hust.edu.cn}
}

\begin{document}
\maketitle
\begin{abstract}
Detecting offensive language is a challenging task. Generalizing across different cultures and languages becomes even more challenging: besides lexical, syntactic and semantic differences, pragmatic aspects such as cultural norms and sensitivities, which are particularly relevant in this context, vary greatly. In this paper, we target Chinese offensive language detection and aim to investigate the impact of transfer learning using offensive language detection data from different cultural backgrounds, specifically Korean and English. We find that culture-specific biases in what is considered offensive negatively impact the transferability of language models (LMs) and that LMs trained on diverse cultural data are sensitive to different features in Chinese offensive language detection. In a few-shot learning scenario, however, our study shows promising prospects for non-English offensive language detection with limited resources. Our findings highlight the importance of cross-cultural transfer learning in improving offensive language detection and promoting inclusive digital spaces.

\textbf{\small Warning}: \emph{\small This paper contains content that may be offensive or upsetting.}

\end{abstract}

\begin{table*}[t]
\centering
\scalebox{0.95}{
\begin{tabular}{lcccc}
\toprule
\textbf{Dataset} & \textbf{Language} & \textbf{Train}                                                  & \textbf{Dev}                                                 & \textbf{Test}                                                \\
\midrule
COLD             & Chinese & \begin{tabular}[c]{@{}c@{}}25726\\ \small (12723:13003=0.98)\end{tabular} & \begin{tabular}[c]{@{}c@{}}6431\\ \small (3211:3220=1.00)\end{tabular} & \begin{tabular}[c]{@{}c@{}}5323\\ \small (2107:3216=0.66)\end{tabular} \\
KOLD             & Korean & \begin{tabular}[c]{@{}c@{}}24257\\ \small(12190:12067=1.01)\end{tabular} & \begin{tabular}[c]{@{}c@{}}8086\\ \small(4076:4010=1.02)\end{tabular} & \begin{tabular}[c]{@{}c@{}}8086\\ \small(4044:4022=1.01)\end{tabular} \\
HatEn            & English & \begin{tabular}[c]{@{}c@{}}9000\\ \small(3782:5217=0.72)\end{tabular}    & \begin{tabular}[c]{@{}c@{}}1000\\ \small(427:573=0.75)\end{tabular}   & \begin{tabular}[c]{@{}c@{}}3000\\ \small(2343:657=3.57)\end{tabular}  \\
\midrule
\midrule
Region           && 8449                                                            & 2104                                                         & 2087                                                         \\
Gender           && 6579                                                            & 1657                                                         & 1551                                                         \\
Race             && 10698                                                           & 2670                                                         & 1685                                                         \\
\bottomrule
\end{tabular}}
\caption{Datasets statistics (\textbf{top}) and topic distributions of COLD (\textbf{bottom}). Particularly, statistics of offensive and non-offensive data and the ratio between them are indicated in \textbf{parentheses}.}
\label{tab:data}
\end{table*}

\section{Introduction}\label{sec:introduction}

The proliferation of offensive language and hate speech in online platforms, especially on social media, has significantly increased in recent years~\cite{zampieri-etal-2019-semeval, zampieri-etal-2020-semeval, gao-etal-2020-offensive}. There is a fine line between offensive language and hate speech as few universal definitions exist~\cite{Davidson_Warmsley_Macy_Weber_2017}.
Therefore, hate speech can be classified as a subtype of offensive language. In this paper, we do not differentiate them in detail, and instead, refer to the task of offensive language detection (OLD).

Despite numerous breakthroughs in the development of NLP methods for OLD~\cite{liu-etal-2022-multiple, rusert-etal-2022-robustness}, some significant obstacles remain unsolved~\cite{vidgen-etal-2019-challenges}, including the shortage of data resources for research purposes and bias in human annotation. Since most of the available approaches and resources for OLD are designed for English~\cite{arango-monnar-etal-2022-resources}, the resulting trained models operate within a monocultural background that caters to English speakers.\footnote{Importantly, ``culture'' is multifaceted and complex. When referring to English speakers, we assume that there are general unique features that characterize them, but of course there is enormous diversity within speakers of the same language. As a first step towards the analysis of cross-cultural OLD, we restrict ourselves to the level of language categories.} However, \citet{schmidt-wiegand-2017-survey} believe that OLD has strong cultural implications, unlike other NLP tasks, because an utterance's offensiveness can vary based on an individual's cultural background.




People with different backgrounds react to inputs differently and communicate differently, so their tolerance for the presence of offensive terms, e.g., slur, may differ, as well as what is altogether considered offensive~\cite{JayJanschewitz2008}. Cultural differences have been explored in humor perception~\cite{10.3389/fpsyg.2019.00123}, swearing reception~\cite{doi:10.1080/0907676X.2021.1913199}, translation in semantic inconsistencies~\cite{doi:10.1177/0022022194254006} and honorifics expression~\cite{song-2015-representing, liu-kobayashi-2022-construction}. Even in less obvious cases, however, they bear meaningful significance on how to pose and solve NLP tasks, as cultures differ with respect to style, values, common ground and topics of interest~\cite{hershcovich-etal-2022-challenges}. 


Therefore, we argue that there is a need for addressing cross-cultural aspects in offensive language detection. Although culture is intricate and challenging to define clearly, language still remains as one of the most straightforward manifestation of culture. While recent work~\cite{ringel-etal-2019-cross, Ranasinghe2021} has demonstrated the effectiveness of cross-lingual transfer learning in the text classification and offensive Language (hate speech) detection, they don't consider the impact of cultural background differences (e.g., Eastern and Western culture).
In this paper, we take a step forward in this direction and explore the influence of offensive content from diverse cultural background on OLD, focusing on evaluation in Chinese.




Our contributions are as follows: 1) We explore the impact of transfer learning using offensive language data from different cultural backgrounds on Chinese offensive language detection (\S\ref{sec:method}).
2) We find cultural differences in offensive language are expressed in the text topics,
and that LMs are sensitive to these differences, learning culture-specific biases that negatively impact their transfer ability (\S\ref{sec: evaluation}). 3) We find that in the few-shot scenario, even with very limited Chinese examples, the model quickly adapts to the target culture.

\section{Related work}\label{sec:relate}

\paragraph{Offensive language detection.}
Although most of the research on OLD has focused on English~\cite{10.1145/3232676}, there exist datasets in multiple languages:
Chinese~\cite{deng-etal-2022-cold},
Korean~\cite{jeong-etal-2022-kold},
Danish~\cite{sigurbergsson-derczynski-2020-offensive}, 
Bengali~\cite{das-etal-2022-hate-speech}, and Nepali~\cite{niraula-etal-2021-offensive}, to name a few. 
However, 
language models commonly rely on prior distributions from training data, that reflects a discourse that is temporally and culturally situated \cite{https://doi.org/10.48550/arxiv.2104.06999}. 
In a comprehensive analysis of geographically-related content and its influence on performance disparities of offensive language detection models, \citet{lwowski-etal-2022-measuring} find that current models do not generalize across locations.
\citet{sap-etal-2022-annotators} call for contextualizing offensive (toxicity) labels in social variables as determining what is toxic is subjective, and annotator beliefs can be reflected in the data collected.



\paragraph{Cross-lingual transfer learning.}
Cross-lingual transfer appears as a potential solution to the issue of language-specific resource scarcity~\cite{lamprinidis-etal-2021-universal}. 
\citet{nozza-2021-exposing} demonstrates the limits of cross-lingual zero-shot transfer for hate speech detection in English, Italian and Spanish.
The benefits of few-shot learning is evident in works from \citet{Stappen2020CrosslingualZA} and 
\citet{rottger-etal-2022-data}, who confirmed the effectiveness of few-shot learning for the task of hate speech detection in under-resourced languages.
\citet{ringel-etal-2019-cross} harness cross-cultural differences for English formality and sarcasm detection based on German and Japanese, respectively. \citet{litvak-etal-2022-offensive} show that, in the context of OLD, knowledge transfer is not bidirectional and efficient transfer learning holds from Arabic to Hebrew in terms of recall. 




\section{Method}\label{sec:method}

\subsection{Datasets}

To explore the influence of different cultural backgrounds on Chinese OLD, the most straightforward approach is 
to adopt OLD datasets whose context and annotation process reflect diverse cultural backgrounds. 
We first select COLD~\cite{deng-etal-2022-cold}, a Chinese benchmark dataset covering the topics of racial, gender, and regional bias as our test dataset. 
We then select two other datasets that will be used in different training scenarios (see \S~\ref{sec:learning_settings}): KOLD~\cite{jeong-etal-2022-kold}, a Korean dataset suited for OLD  covering topics such as race, gender, political affiliation and religion; and HatEn, the English subset of HatEval~\cite{basile-etal-2019-semeval} composed of tweets which tends to capture a Western cultural background. 
Table~\ref{tab:data} reports the statistics of the three datasets and the topic distributions of COLD. Notably, the three languages come from three different language families, making linguistic similarities between them less likely to be a factor in effective transfer learning between the datasets.

\begin{figure*}[t]
    \centering
    \includegraphics[width=1.01\textwidth]{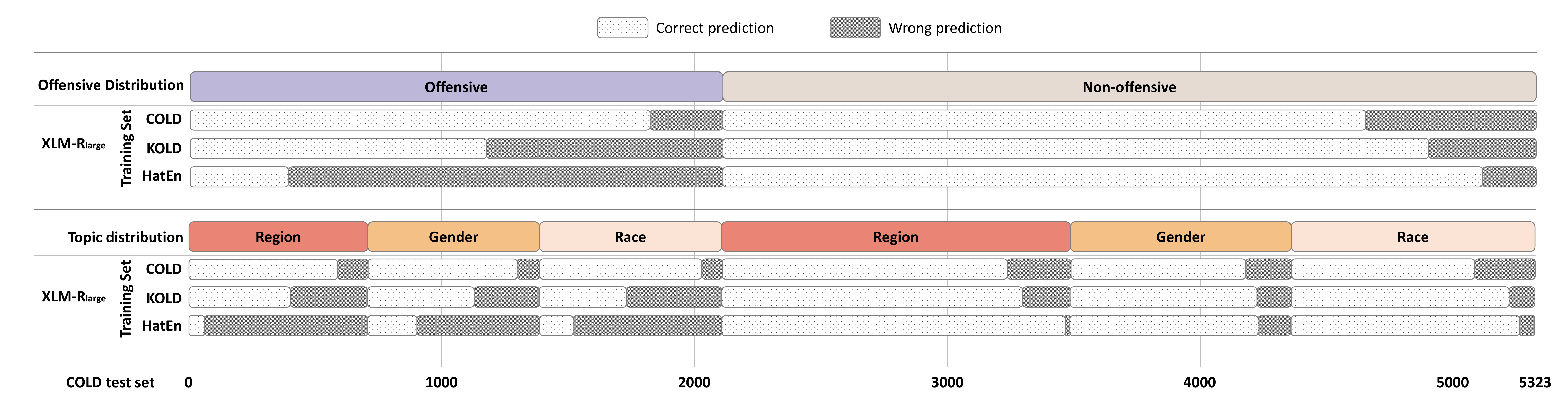}
    \caption{A fine-grained view of the distribution of offensive detection results based on \texttt{XLM-R$_{\texttt{large}}$}. For reference, the colored part represent the distribution of related data in COLD test set. The model learns culture-specific biases---e.g., when training on English, it tends not to classify region-related text as offensive.}
    \label{fig:distribution}
\end{figure*}

\subsection{Learning settings}\label{sec:learning_settings}
We explore different learning settings by utilizing \textbf{intra-cultural} and \textbf{cross-cultural} training sets during fine-tuning. For the intra-cultural setting, we only use COLD as the training set, which ensures cultural consistency in the training and testing process. In the cross-cultural setting, we further set up two ways: 1) \textit{zero-shot}: only use KOLD or HatEn as the training set, which makes the fine-tuning process of LMs come from completely different cultural backgrounds; 2) \textit{mix-training few-shot}: mix COLD with another language (KOLD or HatEn) as the final training set, which introduces cultural interference and makes the acquisition of the target culture more challenging.
For convenience, 
we use $\mathcal{D}\left[X\right]$ to represent the detector with $X$ as training set. Since the datasets are in different languages, we apply multilingual LMs in these experiments.

\paragraph{Translated data setting.}
As an additional control experiment, to avoid the difference from the language itself, we also translate COLD and KOLD into English with \textit{googletrans}\footnote{\url{https://pypi.org/project/googletrans/}} and conduct experiments with \textit{English} PLMs under the same settings.

\section{Experiments}\label{sec: evaluation}
\paragraph{Implementation.}
In our experiments, we only evaluate on COLD and 
try different training settings with COLD, KOLD and HatEn.
In particular, because the data volume of HatEn is relatively small, we use all of its data as the training set. The actual training set of three datasets has offensive data to non-offensive data ratios of 0.98, 1.01, and 1.02 (refer to  Table~\ref{tab:data}). In the cross-cultural zero-shot setting, we also randomly sample 13,000 examples\footnote{The ratio of offensive data to non-offensive data is 0.96.} from the Korean training set to ensure the consistency of the training data sizes with HatEn. For the multilingual LMs, we choose \texttt{mBERT$_{\texttt{base}}$}~\cite{devlin-etal-2019-bert}, \texttt{XLM-R$_{\texttt{base}}$} and \texttt{XLM-R$_{\texttt{large}}$}~\cite{conneau-etal-2020-unsupervised}. 
In the translated data setting, we apply the English models \texttt{BERT$_{\texttt{base}}$}~\cite{devlin-etal-2019-bert}, \texttt{RoBERTa$_{\texttt{base}}$} and \texttt{RoBERTa$_{\texttt{large}}$}~\cite{liu2019roberta}.

Our models are optimized with a learning rate of $5e-5$. We fine-tune each model for 100 epochs using early-stopping with a patience of 5, and run 5 times with different random seeds for each setting.

\paragraph{Overall results.}
\begin{table}[t]
\centering
\scalebox{0.82}{
\begin{tabular}{@{}llll@{}}
\toprule
\textbf{Model}                     & \textbf{Train Set} & \textbf{Test F1} & \textbf{Test ACC} \\ \midrule
\multirow{5}{*}{mBERT$_{\texttt{base}}$}  & COLD        & 77.90±0.25         & 80.86±0.26            \\
                                   & CO+KO                & 78.23±0.05$^{\ast}$         & 81.16±0.19            \\
                                   & CO+HE                & 78.19±0.18$^{\ast}$         & 81.07±0.10                   \\
                                   & KOLD               & 49.27±4.04$^{\ast\ast}$         & 67.85±0.70$^{\ast\ast}$            \\
                                   & KOLD$^{\dagger}$   & 50.34±3.49$^{\ast\ast}$         & 69.47±0.71$^{\ast\ast}$            \\
                                   & HatEn              & 35.96±3.95$^{\ast\ast}$         & 63.54±0.54$^{\ast\ast}$     \\           
\midrule
\multirow{5}{*}{XLM-R$_{\texttt{base}}$}  & COLD        & 78.77±0.27        & 81.51±0.20           \\
                                   & CO+KO                & 78.90±0.10        & 81.78±0.15$^{\ast}$            \\
                                   & CO+HE                & 78.96±0.15        & 81.66±0.18                   \\
                                   & KOLD               & 58.13±1.78$^{\ast\ast}$        & 72.14±0.67$^{\ast\ast}$            \\
                                   & KOLD$^{\dagger}$   & 60.86±1.44$^{\ast\ast}$        & 72.93±0.37$^{\ast\ast}$             \\
                                   & HatEn              & 29.84±2.07$^{\ast\ast}$        & 63.36±0.90$^{\ast\ast}$  \\           
\midrule
\multirow{5}{*}{XLM-R$_{\texttt{large}}$} & COLD               & 79.09±0.24        & 81.87±0.16       \\
                                   & CO+KO                & 79.76±0.19$^{\ast\ast}$        & 82.45±0.19$^{\ast\ast}$            \\
                                   & CO+HE                & 79.43±0.22$^{\ast}$        & 82.16±0.26$^{\ast\ast}$                   \\
                                   & KOLD               & 63.48±1.63$^{\ast\ast}$        & 74.45±0.34$^{\ast\ast}$            \\
                                   & KOLD$^{\dagger}$   & 61.71±2.37$^{\ast\ast}$         & 74.09±0.80$^{\ast\ast}$            \\
                                   & HatEn              & 28.94±2.50$^{\ast\ast}$       & 63.76±0.40$^{\ast\ast}$  \\
\bottomrule
\end{tabular}}
\caption{Overall results on COLD test set. ${\dagger}$ marks KOLD training set is the same size as HatEn. CO, KO and HE are short for COLD, KOLD and HatEn respectively. By conducting Paired Student's t-test, ${\ast}$ = differs significantly from intra-cultural at $p < 0.05$, ${\ast\ast}$ = significant difference at $p < 0.01$.}
\label{tab:results}
\end{table}
The experimental results on COLD test set are shown in Table~\ref{tab:results}.\footnote{We only report the test set score, because only the test set of COLD is annotated manually, and the training and dev sets are labeled semi-automatically.} Compared to the intra-cultural setting, we find that: 1) In the cross-cultural few-shot scenario, the performance differences between $\mathcal{D} \left[\mathrm{COLD}\right]$ and $\mathcal{D}\left[\mathrm{CO+KO}\right]$, $\mathcal{D}\left[\mathrm{COLD}\right]$ and $\mathcal{D}\left[\mathrm{CO+HE}\right]$ are both very small (less than one point at the maximum), which implies that with sufficient knowledge of the Chinese target culture, the intervention of other cultures does not diminish the ability to detect Chinese offensive language, but has a slight contribution. 2) In the cross-cultural zero-shot scenario, the detection ability of $\mathcal{D}\left[\mathrm{KOLD}\right]$ and $\mathcal{D}\left[\mathrm{HatEn}\right]$ get worse. In particular, the former is slightly better than the latter. This implies that it is easier to detect Chinese offensive language in Korean cultural background compared to a Western cultural background.

To better understand the detection ability of Chinese offensive language with different cultural backgrounds, we look closer at offensive detection results for the intra-cultural and cross-cultural zero-shot settings. Figure~\ref{fig:distribution} shows the distribution of the data and the predictions from our best performing model \texttt{XLM-R$_{\texttt{large}}$}. 
First, $\mathcal{D}\left[\mathrm{COLD}\right]$, which is in the same cultural background as the test set, has the best ability to detect offense. 
$\mathcal{D}\left[\mathrm{HatEn}\right]$ is the worst detector, with less than 50\% accuracy for offensive data. Because of this, it can be highly accurate in non-offensive data. This is why $\mathcal{D}\left[\mathrm{HatEn}\right]$ gets a spurious high accuracy on the test set but a very low F1 score (Table~\ref{tab:results}). However, it is noteworthy that the HatEn-trained model requires more severe language to be labeled as offensive,\footnote{This could be a reason to treat Hate Speech Detection as a separate task, contrary to our simplified view here.} so some instances that should be classified as offensive, may not be considered hate speech and will not be classified as such. Moreover, for specific-topic offensive language detection, the performance of each detector is also different, with $\mathcal{D}\left[\mathrm{HatEn}\right]$ performing the worst in the regional topic.

\paragraph{Translated results.}
For the experiments of the translated version of the Chinese and Korean datasets into English. The experimental results are shown in Table~\ref{tab:translated}, showing similar trends to the results in Table~\ref{tab:results}. This demonstrates that the results hold for cross-cultural transfer and are not simply due to linguistic similarities.
\begin{table}[t]
\centering
\scalebox{0.82}{
\begin{tabular}{@{}llll@{}}
\toprule
\textbf{Model}                     & \textbf{Train Set} & \textbf{Test F1} & \textbf{Test ACC} \\ \midrule
\multirow{5}{*}{BERT$_{\texttt{base}}$}  & COLD        & 77.59±0.41         & 80.67±0.37            \\
                                   & CO+KO                & 77.86±0.19$^{\ast}$         & 80.90±0.20            \\
                                   & CO+HE                & 77.50±0.17$^{\ast}$         & 80.47±0.18                   \\
                                   & KOLD               & 61.84±1.46$^{\ast\ast}$      & 71.26±0.34$^{\ast\ast}$            \\
                                   & KOLD$^{\dagger}$   & 61.64±1.06$^{\ast\ast}$      & 71.21±0.27$^{\ast\ast}$            \\
                                   & HatEn              & 21.20±1.36$^{\ast\ast}$       & 61.53±0.21$^{\ast\ast}$     \\           
\midrule
\multirow{5}{*}{RoBERTa$_{\texttt{base}}$}  & COLD   & 77.89±0.46     & 81.01±0.40           \\
                                   & CO+KO                & 78.25±0.40        & 81.35±0.37$^{\ast}$            \\
                                   & CO+HE                & 78.08±0.34        & 81.12±0.25                   \\
                                   & KOLD               & 63.85±1.12$^{\ast\ast}$        & 73.60±0.43$^{\ast\ast}$            \\
                                   & KOLD$^{\dagger}$   & 63.47±0.84$^{\ast\ast}$        & 73.21±0.25$^{\ast\ast}$             \\
                                   & HatEn              & 26.09±2.82$^{\ast\ast}$        & 62.81±0.36$^{\ast\ast}$  \\           
\midrule
\multirow{5}{*}{RoBERTa$_{\texttt{large}}$} & COLD               & 78.22±0.40        & 81.24±0.33       \\
                                   & CO+KO                & 78.74±0.21$^{\ast\ast}$      & 81.70±0.15$^{\ast\ast}$            \\
                                   & CO+HE                & 78.24±0.30$^{\ast}$        & 81.17±0.25$^{\ast\ast}$                   \\
                                   & KOLD               & 65.56±1.16$^{\ast\ast}$        & 73.70±0.49$^{\ast\ast}$            \\
                                   & KOLD$^{\dagger}$   & 64.39±1.60$^{\ast\ast}$         & 73.71±0.37$^{\ast\ast}$            \\
                                   & HatEn              & 26.69±1.38$^{\ast\ast}$       & 63.20±0.44$^{\ast\ast}$  \\
\bottomrule
\end{tabular}}
\caption{The experimental results on the COLD test set, with all training and testing data translated to English. ${\dagger}$ marks KOLD training set is the same size as HatEn. By conducting Paired Student's t-test, ${\ast}$ = differs significantly from intra-cultural at $p < 0.05$, ${\ast\ast}$ = significant difference at $p < 0.01$.}
\label{tab:translated}
\end{table}

\begin{figure}[t!]
    \centering
    \includegraphics[width=\columnwidth]{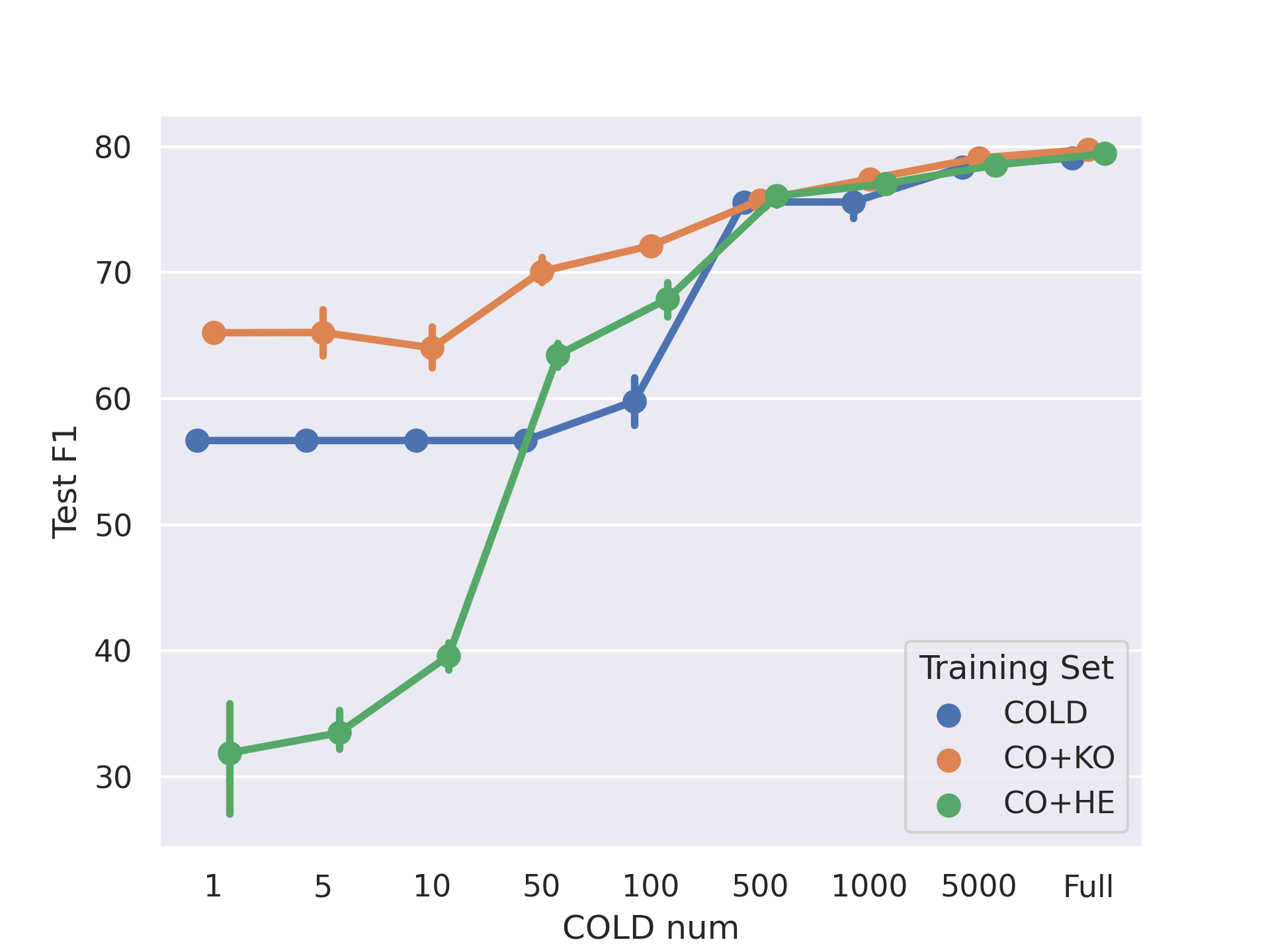}
    \caption{The experimental results (F1) in few-shot setting based on \texttt{XLM-R$_{\texttt{large}}$}, evaluated on the COLD (Chinese) test set. Performance improves rapidly with training examples from the target culture. Pre-training on KOLD (Korean) provides a better starting point, while pre-training on HatEn (English) is detrimental.}
    \label{fig:fewshot}
\end{figure}

\paragraph{Few-shot learning.}
While the diverse cultural backgrounds of Korean and English may not enable precise detection of Chinese offensive language in a zero-shot scenario, it is not detrimental when integrated into the target culture in a few-shot scenario.
Therefore, when mixing heterogeneous cultural background knowledge, is it necessary to provide sufficient target cultural background knowledge?
To investigate this problem, we conduct an analytical experiment under a few-shot setting by incorporating different scales of COLD data into the training set.
Figure~\ref{fig:fewshot} displays experimental results indicating that the correlation between the ability to detect offensive language and target cultural knowledge follows a pattern similar to that of an increasing logarithmic function. 
This implies that offensive language detection performance improves rapidly with limited target cultural knowledge acquisition, but gradually slows down as the amount of target knowledge increases. 
Specifically, when the training focuses on COLD within the range of 1 to 50, $\mathcal{D}\left[\mathrm{COLD}\right]$ possesses limited knowledge of the training concentration, and its detection capability stems primarily from the pre-training model itself. At this stage, HatEn has a clearly negative effect, while KOLD has a positive effect. 
Within the range of 50 to 500, both HatEn and KOLD have an obvious positive effect, while for COLD data scales greater than 500, the effect is still present but less pronounced.
These findings offer promising opportunities for low-resource offensive language detection systems.

\paragraph{Case study.}
To provide an intuitive explanation of cultural differences, we use semantic similarity retrieval \cite{reimers-gurevych-2019-sentence} to find the most similar cases from KOLD to COLD
with the similarity threshold set to 0.7. As depicted in Table \ref{tb:prompt_template}, sentences with similar topics and semantics (e.g. racial discrimination, politics) hold different labels among languages, 
suggesting 
the presence of cultural distinctions in offensive language detection and highlighting the significant obstacles for few-shot learning. Thus, we emphasize the necessity of greater cultural adaptation models that can integrate diverse cultural knowledge.

\CJK{UTF8}{gbsn}
\begin{table}[t]
\centering
\resizebox{\columnwidth}{!}{
    \begin{tabular}{l| l | l}
        \toprule
        \textbf{Chinese} & \textbf{Korean}  & \textbf{Labels}   \\ \hline \hline
        \chinesecolor{黑人反对歧视黑人有啥错？} & \koreancolor{\korean{흑인 대통령도 나온 미국, }} & 0 / 1  \\
        \emph{What is wrong with blacks} & \koreancolor{\korean{이제 인종차별은 사라졌다?}} \\
        \emph{against discrimination} & \emph{America with a black president, } \\ 
        \emph{against blacks?} & \emph{now racism has disappeared?} \\\hline
        \chinesecolor{中国哪有那么容易搞到} & \koreancolor{\korean{중국에서   범은 잡히면 뭐가 }} & 0 / 1 \\
        \chinesecolor{毒品?} & \koreancolor{\korean{잘릴까..}} & \\
        \emph{How can it be so easy} & \emph{What will happen if a }  \\
        \emph{to get drugs in China.}  & \emph{criminal is caught in China?} \\
        \bottomrule
        \end{tabular}}
	\caption{\label{tb:prompt_template} Cases with reversed labels through semantic vector retrieval were listed, suggesting the existence of cultural differences across languages. Non-offensive and offensive cases are labeled as 0 and 1.}
\end{table}

\section{Conclusion}
Our study highlights the challenges of detecting offensive language across different cultures and languages. We show that transfer learning using data from diverse cultural backgrounds have different negative effects on the transferability of language models due to culture-specific biases. However, our findings also indicate promising prospects for improving offensive language detection in promoting inclusive digital spaces, particularly in a few-shot learning scenario.
We call for more research on cross-cultural offensive language detection, which is important to deploy effective moderation strategies for social media platforms, improving cross-cultural communication, and reducing harmful online behavior. 

\section*{Limitations}
Our study explores the impact of transfer learning on offensive language detection using data from different cultural backgrounds. However, treating HatEn as representative of ``Western cultural backgroun'' is too vague, as it ignores the cultural differences between American and British cultures. Moreover, ``culture'' is multifaceted and complex, and there is enormous diversity among speakers of the same language. To focus on language categories, we limit our analysis to a first step towards cross-cultural offensive language detection.

\section*{Ethics Statement}
The datasets used in this study are publicly available, and we strictly follow  the ethical implications of previous research related to the data sources. It is important to note that the content of these datasets does not represent our  opinions or views.

\section*{Acknowledgments}
Thanks to the anonymous reviewers for their helpful feedback. The authors gratefully acknowledge financial support from China Scholarship Council. (CSC No. 202206070002 and No. 202206160052).
\bibliography{anthology,custom}
\bibliographystyle{acl_natbib}

\end{document}